\documentclass[journal]{IEEEtai}

\usepackage{amsmath}
\usepackage{graphicx}
\usepackage{txfonts}
\usepackage{multirow}
\usepackage{verbatim}
\usepackage{array}
\usepackage{subfig}
\usepackage[nocompress]{cite}
\usepackage[colorlinks, linkcolor=red, urlcolor = blue, citecolor=red]{hyperref}

\hypersetup{colorlinks=true,citecolor = blue}

\usepackage{multirow}  
\newtheorem{definition}{Definition}   
\newtheorem{Strategy}{Strategy}
\newtheorem{Property}{Property}

\usepackage{amsmath}
\usepackage{arydshln} 
\usepackage{color} 

\usepackage{stfloats} 

\usepackage[ruled,linesnumbered]{algorithm2e}

\ifCLASSINFOpdf

\else

\fi

\begin{document}

\title{Itemset Utility Maximization with Correlation Measure}

\author{Jiahui Chen, Yixin Xu, Shicheng Wan, Wensheng Gan*, and Jerry Chun-Wei Lin,~\IEEEmembership{Senior Member,~IEEE}

\thanks{This research was supported in part by the National Natural Science Foundation of China (Grant Nos. 61902079 and 62002136), Guangzhou Basic and Applied Basic Research Foundation (Grant Nos. 202102020928 and 202102020277), National Key-Research and Development Program of China (Grant No. 2020YFB2104003), and the Young Scholar Program of Pazhou Lab (Grant No. PZL2021KF0023).}
	
\thanks{Jiahui Chen, Yixin Xu, and Shicheng Wan are with the School of Computer Science and Technology, Guangdong University of Technology, Guangzhou 510006, China. (E-mail: csjhchen@gmail.com, yxxu1998@gmail.com, and scwan1998@gmail.com)}

\thanks{Wensheng Gan is with the College of Cyber Security, Jinan University, Guangzhou 510632, China; and also with Pazhou Lab, Guangzhou 510330, China. (E-mail: wsgan001@gmain.com)}

\thanks{Jerry Chun-Wei Lin is with the Department of Computer Science, Electrical Engineering and Mathematical Sciences, Western Norway University of Applied Sciences, Bergen, Norway. (E-mail: jerrylin@ieee.org)}

\thanks{Corresponding author: Wensheng Gan}
}

\maketitle

\begin{abstract}
	As an important data mining technology, high utility itemset mining (HUIM) is used to find out interesting but hidden information (e.g., profit and risk). HUIM has been widely applied in many application scenarios, such as market analysis, medical detection, and web click stream analysis. However, most previous HUIM approaches often ignore the relationship between items in an itemset. Therefore, many irrelevant combinations (e.g., \{gold, apple\} and \{notebook, book\}) are discovered in HUIM. To address this limitation, many algorithms have been proposed to mine correlated high utility itemsets (CoHUIs). In this paper, we propose a novel algorithm called the \underline{I}temset \underline{U}tility \underline{M}aximization with \underline{Co}rrelation Measure (CoIUM), which considers both a strong correlation and the profitable values of the items. Besides, the novel algorithm adopts a database projection mechanism to reduce the cost of database scanning. Moreover, two upper bounds and four pruning strategies are utilized to effectively prune the search space. And a concise array-based structure named utility-bin is used to calculate and store the adopted upper bounds in linear time and space. Finally, extensive experimental results on dense and sparse datasets demonstrate that CoIUM significantly outperforms the state-of-the-art algorithms in terms of runtime and memory consumption.
\end{abstract}

\begin{IEEEImpStatement}
	This article contributes to a utility-based correlated pattern discovery method for transaction databases. It studies the relationship between consisted items of a profitable itemset. Compared with the state-of-the-art algorithms, the designed CoIUM method has greatly improved the performance in terms of execution time and memory consumption. The proposed method is good at dealing with massive database, especially dense databases, which is very beneficial to dealing with commodity sales in real life. Based on large-scale testing, it is found that this method has good scalability in terms of runtime and memory consumption, which indicates that CoIUM is suitable for different types of databases (include sparse and dense). Since the traditional high utility itemset mining has been successfully applied in many applications and various domains, this correlated utility mining problem formulation and efficient algorithm can be applied in many applications, such as product recommendation, market basket analysis, e-learning, text mining and web click stream analysis.
\end{IEEEImpStatement}


\IEEEpeerreviewmaketitle


\section{Introduction}


Data mining techniques (e.g., pattern mining \cite{fournier2022pattern,gan2019survey}) play a vital role in many real-life applications, such as market analysis \cite{aggarwal2014frequent} and website click stream analysis \cite{chu2008efficient}. In previous studies on data mining, a fundamental task called frequent itemset mining (FIM) \cite{agrawal1994fast,han2004mining,luna2019frequent} is used to discover itemsets that frequently occur in a transaction database. FP-Growth \cite{han2004mining} is one of the most famous FIM algorithms. It maintains the original database in an FP-tree structure \cite{han2004mining}. However, the limitation of traditional FIM algorithms is that different items have identical importance (e.g., unit profit or weight). For instance, the profit of a piece of bread is equal to that of a diamond in the FIM domain. This assumption is unrealistic in real business life. 

To address the above limitations, a new framework known as high utility itemset mining (HUIM) \cite{ahmed2009efficient,gan2021survey,liu2012mining} was proposed. Compared to FIM, HUIM considers both the quantity and profit of an item. An itemset is referred to as a high utility itemset (HUI) if its utility value is higher than or equal to the user-specified \textit{minUtil}. HUIM has attracted considerable attention in recent decades because discovering profitable patterns is more valuable than discovering frequent patterns for enterprises. In FIM, the support of an item has an anti-monotonic property, which means that the support of an itemset is not less than that of its supersets. This property can efficiently reduce the search space because supersets of infrequent itemsets are not considered. However, the utility of an item in HUIM does not hold this property. For example, in a supermarket, assume that the amount of the daily sales of \{\textit{bread}\} is \$100, but the amount of daily sales of its supersets (i.e., \{\textit{bread},\textit{milk}\}) may be less than, equal to, or higher than \$100. This indicates that the utility is neither monotonic nor anti-monotonic. To address this issue, transaction-weighted utilization (abbreviated as \textit{TWU}), which has the anti-monotonic property, was proposed by Liu \textit{et al.} \cite{liu2005two}. Based on the \textit{TWU} measure, HUIM can prune the search space effectively. After that, a lot of HUIM algorithms \cite{liu2012mining, gan2018coupm, tseng2010up} adopted \textit{TWU} to improve their performance. In general, HUIM approaches can be summarized into two types: two-phase and single-phase algorithms. The two-phase algorithms are characterized by discovering HUIs in two steps. In the first step, numerous candidates are generated by overestimating the utility of itemsets. Then, they calculate the real utility of candidates to filter out HUIs. These two-phase algorithms can discover a complete set of HUIs, there are two shortcomings that cannot be ignored: 1) generating many candidates in the first phase; and 2) repeatedly scanning the database. As a result, two-phase algorithms suffer from long execution times and memory consumption. To solve these issues, the first single-phase HUIM algorithm called HUI-Miner \cite{liu2012mining} was proposed. Compared to the two-phase algorithms, the single-phase algorithms do not need to repeatedly scan the database. The existing single-phase algorithms include: FHM \cite{fournier2014fhm}, EFIM \cite{zida2017efim}, MEFIM \cite{nguyen2019mining}, etc.

However, a key disadvantage of HUIM algorithms is that they usually ignore the relationships between items. This case will usually reveal numerous irrelevant items in high-utility itemsets. For example, in market analysis, assume that \{\textit{notebook}, \textit{book}\} and \{\textit{phone}, \textit{TV}\} are high utility patterns. Obviously, these products (i.e., \textit{notebook}, \textit{book}, \textit{phone}) are weakly correlated. Instead, in real life, retailers often hope to earn more revenue by making related products into product combinations, such as \{\textit{milk}, \textit{bread}\}, \{\textit{gold}, \textit{diamond}\}, and \{\textit{computer}, \textit{mouse}\}. In essence, these products have a strong connection because they often meet the needs of most customers. To discover more correlated high utility itemsets (CoHUIs), some interesting measures such as \textit{Kulc} \cite{wu2010re}, bond \cite{omiecinski2003alternative} and coherence \cite{barsky2011mining} were proposed as standards. Based on the definition of \textit{Kulc}, many algorithms like CoHUIM \cite{gan2018extracting}, CoUPM \cite{gan2018coupm} and ECoHUPM \cite{saeed2021efficient} were adopted to discover CoHUIs. These algorithms improved the mining performance by using novel data structures and pruning strategies. Recently, Vo \textit{et al.} \cite{vo2020mining} presented the CoHUI-Miner algorithm. They used projection techniques to reduce the expense of database scanning and utilized several pruning strategies to mine CoHUIs effectively. However, CoHUI-Miner only used loose upper bound (\textit{TWU} and remaining utility) to prune the search space, which led to numerous unpromising candidates being generated. Therefore, it suffered in terms of runtime and memory consumption.

To this end, we propose an efficient projection-based algorithm, namely \textbf{I}temset \textbf{U}tility \textbf{M}aximization with \textbf{Co}rrelation Measure (CoIUM). Compared to other state-of-the-art algorithms, CoIUM can discover a complete set of CoHUIs more effectively with less execution time and memory consumption. The main contributions of our work are as follows:

\begin{itemize}
	\item We put forward a single-phase algorithm which utilizes two tighter upper-bounds to discover all CoHUIs without candidate generation. Also, the utility-bin structure \cite{zida2017efim} is adopted to calculate and store the upper bounds of itemsets.
	
	\item The CoIUM algorithm uses a projection technique that can effectively prune the search space.
	
	\item The CoIUM algorithm adopts a measure called $Kulc$ as the correlation factor. It is used as a metric to distinguish whether items are highly correlated.
	
	\item Massive experiments are conducted on several datasets to evaluate the performance of CoIUM. We also compare the performance of CoIUM with the state-of-the-art algorithms. The results show that the CoIUM is the most efficient in terms of runtime and memory consumption.
\end{itemize}

The rest of this paper is organized as follows. Some related works about HUIM and CoHUIM are introduced in Section \ref{sec:relatedwork}, respectively. In Section \ref{sec:preliminaries}, we introduce the preliminaries and problem statement of correlated high-utility itemset mining. In addition, a novel CoIUM algorithm is proposed in Section \ref{sec:algorithm}. Several experimental results are presented in Section \ref{sec:experiments}. Finally, the conclusion and future work are presented in Section \ref{sec:conclusion}.

\section{Related Work}
\label{sec:relatedwork}


\subsection{High Utility Itemset Mining}

A key difference between high utility itemset mining (HUIM) \cite{gan2021survey,liu2012mining,qu2020mining,nguyen2021efficient} and frequent itemset mining (FIM) \cite{han2004mining,luna2019frequent} is that frequency has the anti-monotonic property, but utility does not. Hence, if itemset is unpromising, the HUIM cannot easily filter out all its supersets in advance. To address this limitation, transaction-weight utilization (\textit{TWU}), which is anti-monotonic, was first proposed in Two-Phase \cite{liu2005two}. Based on the \textit{TWU} measure, if \textit{TWU} of a subset is lower than a given \textit{minUtil} threshold, all its supersets are not HUIs \cite{liu2005two}. Therefore, almost HUIM algorithms reduce the search space by using the \textit{TWU} measure. However, \textit{TWU} is a loose upper bound for eliminating candidates. Thus, this measure always overestimates the utilities of itemsets, which may result in massive candidates. At the same time, IHUP \cite{ahmed2009efficient} constructed a tree structure to discover HUIs. Compared with Two-Phase, this new structure can save more runtime and memory. Thus, the experimental results showed that IHUP was about three times faster than Two-Phase in runtime. However, the upper bound of IHUP was not tight, which may have retained numerous unpromising candidates. To reduce the generation of unpromising candidates and identify HUIs effectively, Tseng \textit{et al.} \cite{tseng2010up} proposed UP-Growth. It also constructed a pattern tree to extract HUIs and utilized several novel strategies to discover HUIs effectively. UP-Growth can reduce the numerous unpromising candidates and enhance mining performance, so it outperformed IHUP in terms of runtime and memory.

The above-mentioned algorithms are two-phase algorithms, and they face the problems of generating numerous unpromising candidates and scanning the database at least twice. To address these shortcomings, a single-phase approach was proposed to discover HUIs. As a single-phase approach, HUI-Miner \cite{liu2012mining} algorithm utilized a new data structure called utility-list, which can retain key information about HUIs without scanning the database multiple times. HUI-Miner used a depth-first search to discover all HUIs without omission, and constructed the utility-list of each itemset by joining the utility-lists of its subsets. HUI-Miner can reduce numerous candidates because it uses the remaining utility, which is more compact than \textit{TWU}. As for performance, HUI-Miner is superior to all two-phase algorithms. Meanwhile, another algorithm called $d^2$HUP was presented by Liu \textit{et al.} \cite{liu2012direct}. HUI-Miner and $d^2$HUP were proposed at the same time, but their key pruning strategies are conceptually equivalent. However, using join operations to create utility-lists is time- and memory-consuming. To reduce the expense of join operations, the FHM algorithm \cite{fournier2014fhm} used the EUCP strategy to effectively reduce the generation of join operations. The experiments showed that the performance of FHM is up to 6 times faster than that of HUI-Miner. Then, an algorithm called IMHUP was proposed \cite{ryang2017indexed}, and utilized an indexed list-based structure to discover HUIs without requiring numerous comparison operations. Based on this structure, two techniques were developed to improve mining performance. In addition, ULB-Miner \cite{duong2018efficient} utilized a utility-list buffer structure to store information and introduced a new index segment to access the stored data. Thus, the performance of ULB-Miner is better than that of HUI-Miner and FHM. EFIM \cite{zida2017efim} utilized two tighter upper bounds to discover the HUIs effectively. Furthermore, it utilized two new techniques (projection technique and transaction merging) to deal with long transactions. The experiments showed that the performance of EFIM is faster than that of HUI-Miner and FHM on spare and dense databases. The above algorithms are devised under the condition that each item in the database has a different positive utility. Then, the FHN algorithm \cite{lin2016fhn} can discover the HUIs with negative unit profit, and Gan \textit{et al.} \cite{gan2020tophui} proposed a top-$k$ HUIM algorithm to solve the threshold setting issue. To summarize, many utility mining algorithms have been developed to solve practical problems, including on-shelf utility mining \cite{chen2022shelf}, and high average-utility itemset mining \cite{song2021generalized,wan2022discovering}. Up until now, a lot of HUIM algorithms with various data have been developed, such as high utility association rules \cite{mai2017lattice}, HUIM with fuzzy itemsets \cite{wan2021fuim}, targeted utility querying \cite{miao2022targeted}, HUIM on massive data \cite{han2021efficient}, HUIM on noisy data \cite{baek2021approximate}, and HUIM on stream data \cite{song2021topums}.


\subsection{Correlated High Utility Itemset Mining}

Most HUIM algorithms aim to discover a complete set of HUIs, but do not consider the relationships between items in an itemset. To find potential and strong affinity patterns, WIP \cite{yun2007efficient} was proposed, which used a weighted confidence measure to generate strong affinity patterns. Besides, there are three other interesting measures for correlation: any-confidence, all-confidence, and bond \cite{ahmed2011framework}. These measures are useful in discovering the degree of connection between items. However, these measures cannot be properly applied to databases that contain numerous empty transactions. To solve this problem, a null-transaction invariant \cite{wu2010re} measure was proposed. It was a credible measure because it did not consider how many null-transactions appeared in the database. There are five null-(transaction) invariant measures: all-confidence \cite{omiecinski2003alternative}, coherence \cite{omiecinski2003alternative}, cosine \cite{wu2010re}, kulczynsky \cite{wu2010re}, and max-confidence \cite{omiecinski2003alternative}. Compared with previous studies on traditional data mining approaches, the proposed affinity/correlation patterns can mine more useful information efficiently. 

To discover more interesting information, Ahmed \textit{et al.} \cite{ahmed2011framework} first proposed a high-utility itemset mining algorithm to discover HUIs with frequency affinity. The HUIPM algorithm utilized a tree structure UFTA to store HUIs and used a pruning strategy to reduce the generation of unpromising candidates. However, the limitation of HUIPM is that it is time-consuming because it has to establish tree nodes and generate lots of candidates. Thus, Lin \textit{et al.} \cite{lin2017fdhup} presented a single-phase algorithm known as FDHUP which uses two data structures called EI-table and FU-tree, to discover correlated itemsets efficiently. The FDHUP utilized three pruning strategies, which were based on RAU (remaining affinitive utility) to prune the candidates. As a result, FDHUP performed better than HUIPM in terms of runtime and memory. FCHM \cite{fournier2020mining} also discovered the correlated high utility itemsets. It has two versions, FCHM$_{bond}$ depended on bond measure and FCHM$_{\textit{all-confidence}}$ relied on all-confidence measure. The algorithm used two structures, utility-list and EUCS, to calculate the \textit{TWU} values. Then, Gan \textit{et al.} \cite{gan2018extracting} proposed CoHUIM. This algorithm used a projection mechanism to improve the performance of the scanning database and utilized the property of \textit{Kulc} to prune the unpromising candidates. They also developed the CoUPM algorithm \cite{gan2018coupm}, which utilized utility-list structure to store promising itemsets and used \textit{Kulc} to calculate the support of itemsets. The algorithm also utilized some properties of utility and correlation to effectively explore the search space. Then, Rashad \textit{et al.} \cite{saeed2021efficient} proposed the ECoHUPM algorithm to evaluate the correlation of HUIs. This algorithm used a new data structure named CoUTlist to store utility and support for each itemset. It also introduced three new pruning properties to prune unpromising candidates. Recently, Vo \textit{et al.} \cite{vo2020mining} introduced CoHUI-Miner to extract correlated items that provide numerous benefits. This algorithm applied projection technique to reduce the expense of scanning the database and applied the prefix utility of projection technique to compute the utility value. The experiments showed that it was efficient, especially on dense databases.

\section{Preliminaries and Problem Formulation}
\label{sec:preliminaries}

In this section, we present the basic definitions and problem formulation. More definition details can be referred to previous work \cite{vo2020mining,gan2018coupm,zida2017efim}. Let $I$ = \{$i_1$, $i_2$, $\dots$, $i_m$\} be a set of $m$ distinct items in a transaction database $\mathcal{D}$ = \{$T_1$, $T_2$, $\dots$, $T_n$\}, where $n$ is the number of transactions in $\mathcal{D}$. An $k$-itemset is a subset of $I$ which contains $k$ different items. Each transaction consists of some distinct items and has a unique identifier of $T_c$ called \textit{Tid}. Each item $i \in I$ owns an external utility (e.g., unit profit) that is represented by $pr(i)$. And each item $i$ in transaction $T_c$ has an internal utility (e.g., purchase quantity) which denoted as $q(i,T_c)$. In addition, Table \ref{table:database} is a simple transaction database (the third column is transaction utility) which consists of six distinct items (i.e., $A$, $B$, $C$, $D$, $E$, and $F$), and Table \ref{table:profit} shows the external utilities of all items.

\begin{table}[!ht]
    \normalsize
	\centering
	\caption{A transaction database}
	\setlength{\tabcolsep}{3.8mm}{
	\begin{tabular}{ccc} \hline
		\textbf{Tid}   &   \textbf{Transaction}    &   \textbf{Utility}     \\ \hline
		$T_1$ &  ($A$,1) ($B$,2) ($C$,2) ($D$,3)   &  \$19   \\ 
		$T_2$ &  ($A$,2) ($B$,1) ($E$,3) ($F$,2)    &  \$18     \\ 
		$T_3$ &  ($A$,4) ($C$,3) ($E$,2)     &  \$23    \\ 
		$T_4$ &  ($A$,3) ($C$,2) ($F$,3)     &  \$17    \\ 
		$T_5$ &  ($B$,3) ($C$,4) ($D$,5)    &  \$25     \\ 
		$T_6$ &  ($B$,2) ($C$,3) ($D$,3) ($F$,5)    &  \$21     \\ 
		$T_7$ &  ($A$,1) ($B$,2) ($C$,2) ($D$,3) ($E$,5) ($F$,2)    &  \$31     \\ 
		$T_8$ &  ($C$,2) ($D$,3) ($E$,1)    &  \$13     \\ \hline
	\end{tabular}
	}
	\label{table:database}
\end{table}

\begin{definition}[Utility of an itemset]
	\rm In a transaction $T_c$, the utility of an item $i$ is defined as $u(i,T_c)$ = $pr(i)$ $\times$ $q(i,T_c)$, where $i \in T_c$. Similarly, the utility of an itemset $X$ is denoted as $u(X,T_c)$ = $\sum_{i \in X \land X \subseteq T_c}u(X,T_c)$. Furthermore, in database $\mathcal{D}$, the utility of an itemset $X$ is denoted as $u(X)$ = $\sum_{X \subseteq T_c \land T_c \in \mathcal{D}}u(X,T_c)$. 
\end{definition}

For example, the utility of item $A$ in transaction $T_2$ is $u(A,T_2) =$ \$4 $\times$ 2 = \$8. The utility of itemset $\{A,B\}$ in transaction $T_2$ is $u(\{A, B\},T_2)$ = $u(A,T_2)$ + $u(B,T_2)$ = \$8 + \$2 = \$10. At the same time, $u(\{A,B\})$ = $u(\{A, B\},T_1)$ + $u(\{A,B\},T_2)$ + $u(\{A,B\},T_7)$ = \$8 + \$10 + \$8 = \$26.

\begin{table}[!ht]
    \normalsize
	\centering
	\caption{External utility of each item}
	\setlength{\tabcolsep}{4.2mm}{
	\begin{tabular}{ccccccc} \hline
		\textbf{Item} & $A$ & $B$ & $C$ & $D$ & $E$ & $F$  \\ \hline
		\textbf{Utility(\$)} & 4 & 2 & 1 & 3 & 2 & 1   \\ \hline
	\end{tabular}
	}
	\label{table:profit}
\end{table}

\begin{definition}[Utility of database]
	\rm The utility of a transaction $T_c$ is defined as $tu(T_c)$ = $\sum_{i_j \in T_c}u(i_j,T_c)$, where $i_j$ is the $j$-th item in $T_c$. Obviously, the total utility of a database $\mathcal{D}$ is the summation of all transactions it contains, where \textit{TU} = $\sum_{T_c \subseteq \mathcal{D}}tu(T_c)$.
\end{definition}

For example, the utility of $T_1$ is $tu(T_1)$ = $u(A,T_1)$ + $u(B,T_1)$ + $u(C,T_1)$ + $u(D,T_1)$ = \$4 + \$4 + \$2 + \$9 = \$19. Then, we can calculate the utility of other seven transactions with the same manner. And the total utility of database $\mathcal{D}$ is \$167. Based on the transaction-weighted utilization \cite{liu2005two}, we have \textit{TWU}($\{A,B\}$) = $tu(T_1)$ + $tu(T_2)$ + $tu(T_7)$ = \$19 + \$18 + \$31 = \$68. Table \ref{table:TWU} lists \textit{TWU} of all 1-itemsets.

\begin{table}[!ht]
    \normalsize
	\centering
	\caption{Support and \textit{TWU} of all 1-itemsets}
	\setlength{\tabcolsep}{2.6mm}{
	\begin{tabular}{ccccccc} \hline
		\textbf{Item} & $A$ & $B$ & $C$ & $D$ & $E$ & $F$ \\ \hline
		\textbf{TWU}  & \$108 & \$114 & \$149 & \$109 & \$85 & \$87 \\
		\textbf{Support}  & 5 & 5 & 7 & 5 & 4 & 4 \\ \hline
	\end{tabular}
	}
	\label{table:TWU}
\end{table}

There are many measures \cite{ahmed2011framework} used to evaluate the correlation between different items, but some of them will be affected by empty transactions. In this paper, we introduce the Kulczynsky measure \cite{wu2010re} (abbreviated as \textit{Kulc}), which is not affected by null transactions.

\begin{definition}[Kulc]
	\rm The frequency of an itemset $X$ in database $\mathcal{D}$ is the amount of $X$ occurs. We use \textit{sup($X$)} to represent the frequency of $X$. For an $k$-itemset $X$ = $\{i_1,$ $i_2,$ $\dots, i_k\}$, its \textit{Kulc} is defined as follows: \textit{Kulc}($X$) = $\frac{1}{k} \times \sum_{i_j \in X}\frac{\textit{sup}(X)}{\textit{sup}(i_j)}$.
\end{definition}

For example, \textit{Kulc}(\{$A,B$\}) = $\frac{1}{2}$ $\times$ ($\frac{\textit{sup}(\{A,B\})}{\textit{sup}(A)} + \frac{\textit{sup}(\{A,B\})}{\textit{sup}(B)}$) = $\frac{1}{2} \times (\frac{3}{5} + \frac{3}{5})$ = 0.6. Obviously, the range of \textit{Kulc} is [0,1] \cite{gan2018extracting}. In fact, \textit{Kulc} represents the average of frequency, so if \textit{Kulc}($X$) becomes high, related items in $X$ appear frequently. Compared to other correlation measures, \textit{Kulc} holds the \textit{null} (transaction)-invariant property. It is more suitable for applications in different fields because it is independent on database size \cite{gan2018extracting}.

\begin{definition}[Correlated high-utility itemset]
	\rm An itemset $X$ can be assumed as a correlated high utility itemset (CoHUI) when it meets two conditions: (1) $u(X)$ $\geq$ \textit{minUtil} $\times$ \textit{TU}; (2) \textit{Kulc}($X$) $\geq$ \textit{minCor}, where \textit{minCor} is a user-defined minimum correlation threshold.
\end{definition}

For example, in a database, \textit{minUtil} and \textit{minCor} are set to 0.1 and 0.7, respectively. The itemset $\{A,B\}$ is an HUI because the utility of $\{A,B\}$ is not less than \textit{minUtil} $\times$ \textit{TU} ($u(\{A,B\})$ = \$26 $>$ \textit{minUtil} $\times$ \textit{TU} = 0.1 $\times$ \$168 = \$16.8), but it is not a CoHUI since its \textit{Kulc} is less than \textit{minCor} (\textit{Kulc}(\{$A,B$\}) = 0.6 $<$ 0.7).

\textbf{Problem formulation}: In a transaction database, given two user-specified minimum thresholds (\textit{minUtil} and \textit{minCor}), the purpose of CoIUM is to find all CoHUIs that their utilities are greater than \textit{minUtil} $\times$ \textit{TU} and their \textit{Kulc} is higher than \textit{minCor}.

\section{Proposed CoIUM Algorithm}
\label{sec:algorithm}

In this section, we present a single-phase algorithm to extract all CoHUIs. Subsection \ref{sec:alg_upper} discusses several upper bounds and pruning strategies. Subsection \ref{sec:alg_projection} introduces an effective projection technology to reduce the cost of database scanning. Subsection \ref{sec:alg_code} introduces the pseudocode of the CoIUM algorithm. In Subsection \ref{sec:alg_example}, we use a detailed example to discuss how CoIUM works.
 
\subsection{Several Strategies for Pruning Search Space} 
\label{sec:alg_upper}

The proposed CoIUM employs a depth-first method to explore the search space. Given a set of distinct items $I$, then the search space ($2^{\mid I \mid}$) can be represented as a set-enumeration tree \cite{rymon1992search}. The algorithm starts with an empty set. To discover all CoHUIs of itemsets $\alpha$, CoIUM recursively appends its extension to $\alpha$ through the $\succ$ order, and this makes $\alpha$ becomes larger. We assume the items $I$ = \{$A$, $B$, $C$, $D$, $E$, $F$\} and the total order $\succ$ is \textit{TWU} ascendant order for the running example. Thus, the order is \{$E$ $\prec$ $F$ $\prec$ $A$ $\prec$ $D$ $\prec$ $B$ $\prec$ $C$\}.

\begin{Property}
	\label{sec:alg_property1}
	\rm Given an itemset $X$, \textit{TWU} of $X$ is always higher than or equal to its utility (\textit{TWU}($X$) $\geq$ $u(X)$).
\end{Property}

\begin{Strategy}
	\rm Based on Property \ref{sec:alg_property1}, for any itemset $X$, if \textit{TWU}($X$) $<$ \textit{minUtil} $\times$ \textit{TU}, then all supersets of $X$ are low-utility itemsets. Thus, we can remove the low utility itemsets from the search space. Proof details are provided in study \cite{liu2005two}.
\end{Strategy}

\begin{definition}[Extension of an itemset \cite{zida2017efim}]
	\rm Consider an itemset $\alpha$, and $E$ ($\alpha$) is represented as its extension according to the $\succ$ order. E($\alpha$) = \{$z$ $|$ $z$ $\in$ $I$ $\land$ $z$ $\succ$ $i$, $\forall$ $i$ $\in$ $\alpha$\}. Given an $k$-itemset $X$, we denote its extensions $Z$ containing $(k+i)$ items as $i$-extension of $X$.
\end{definition}

For instance, consider the database of our running example and an itemset $X$ = \{$A$\}. The 1-extension of $X$ are \{$A,B$\}, \{$A,C$\}, \{$A,D$\}, and the itemset \{$A,B,C$\} is called a 2-extension of \{$A$\}. To prune more unpromising candidates, it is important to use a more compact upper bound to limit the production of unpromising candidates. We present two upper bounds (\textit{TWU} and \textit{re}) that have been used in previous work, and introduce two more compact upper bounds (\textit{su} and \textit{lu}) used in this work. Using different strategies based on these upper bounds can prune different numbers of candidates. In addition, we introduce the property of \textit{Kulc}.

\begin{definition}[Remaining utility \cite{liu2012mining}]
	\rm  Given an itemset $X$ in transaction $T_c$. We define $\succ$ as a total order on items from $I$. The remaining utility of $X$ in $T_c$ is denoted as $re(X,T_c)$ = $\sum_{i \in T_c \land i \succ x \forall x \in X }u(i,T_c)$.
\end{definition}

For example, we select an itemset $\{A,D\}$ from transaction $T_1$. Based on \textit{TWU} ascendant order, items \{$B$\} and \{$C$\} appear after itemset $\{A,D\}$. The remaining utility of $\{A,D\}$ is $re(\{A,D\},T_1)$ = $u(B,T_1)$ + $u(C,T_1)$ = \$4 + \$2 = \$6.

\begin{definition}[Local utility \cite{zida2017efim} and subtree utility \cite{zida2017efim}]
	\rm  Given an item $z$ and an itemset $\alpha$, where $z \in E(\alpha)$. The local utility of $\alpha$ w.r.t. $z$ is defined as $lu(\alpha,z)$ = $ \sum_{(\alpha \bigcup \lbrace z \rbrace) \subseteq T_c \wedge T_c \subseteq \mathcal{D}}[u(\alpha,T_c) + re(\alpha,T_c)]$. The subtree utility of $\alpha$ and $z$ is defined as $su(\alpha,z)$ = $\sum_{(\alpha \bigcup z) \subseteq T_c \wedge T_c \subseteq D}[u(\alpha,T_c) + u(z,T)]$ + $\sum_{i \in T_c \wedge i \in E(\alpha \bigcup \lbrace z \rbrace)} u(i,T_c)$.
\end{definition}

For example, we select $\alpha$ = \{$A$\}. The local utility of $\alpha$ and extension item $D$ is $lu(\alpha,\{D\})$ = $(u(\{A\},T_1)$ + $re(\{A\},T_1)$ + $u(\{A\},T_7)$ + $re(\{A\},T_7))$ = \$19 + \$19 = \$38.
The subtree utility of $\alpha$ and extension item $C$ is $su(\alpha,\{C\})$ = ($u(\{A\},T_1)$ + $u(\{C\},T_1)$ + \$0) + ($u(\{A\},T_3)$ + $u(\{C\},T_3)$ + \$0) + ($u(\{A\},T_4) $ + $u(\{C\},T_4) $ + \$0) + ($ u(\{A\},T_7) $ + $u(\{C\},T_7) $ + \$0) = [\$4 + \$2] + [\$16 + \$3] + [\$12 + \$2] + [\$4 + \$2] = \$45.



\begin{Property}
	\label{sec:alg_property2}
	\rm Given an item $z$ and itemset $\alpha$, where $z \in E(\alpha)$. We get relationship $lu(\alpha,z) \ge u(Z)$, where $Z$ = $\alpha \cup z$.
\end{Property}

\begin{Strategy}
	\label{sec:alg_Strategy1}
	\rm Based on Property \ref{sec:alg_property2}, given an item $z$ and itemset $\alpha$, where $z \in E(\alpha)$. If $lu(\alpha,z)$ $<$ \textit{minUtil} $\times$ \textit{TU}, all extensions of $\alpha$ containing $z$ are low-utility in a subtree. This means that the item $z$ should be removed in the search space of $\alpha$. Proof details are provided in study \cite{zida2017efim}.
\end{Strategy}

For example, if the algorithm discovers itemset \{$A,B$\} having local utility value is less than \textit{minUtil} $\times $ \textit{TU}, all extensions of \{$A,B$\}, e.g., \{$A,B,C$\} and \{$A,B,D$\}, can be ignored.

\begin{Property}
	\label{sec:alg_property3}
	\rm Given an item $z$ and itemset $\alpha$, where $z \in E(\alpha)$. The relationship $su(\alpha,z) \ge u(\alpha \bigcup \{z\})$ always holds on. It also applies to $Z$ which is an extension of $\alpha \cup \{z\}$, that is $su(\alpha,z) \ge u(Z)$.
\end{Property}

\begin{Strategy}
	\label{sec:alg_Strategy2}
	\rm Based on Property \ref{sec:alg_property3}, given an item $z$ and itemset $\alpha$, where $z \in E(\alpha)$. If $su(\alpha,z)$ $<$ \textit{minUtil} $\times$ \textit{TU}, all extensions of $\alpha$ containing $z$ are low-utility. In the set-enumeration tree, this means that it is unnecessary to explore $\alpha \cup z$ and all of its supersets. Proof details are provided in study \cite{zida2017efim}.
\end{Strategy}

Based on Strategy \ref{sec:alg_Strategy1}, the unpromising itemsets will be filtered out from the subtrees of an item $\alpha$. To reduce the cost of exploring search space, Strategy \ref{sec:alg_Strategy2} is proposed to identify whole subtrees of $\alpha$ which can be extended.


\begin{definition}[Primary and secondary items \cite{zida2017efim}]
	\rm  Given an item $z$ and itemset $\alpha$, where $z \in E(\alpha)$. The $primary$ items of $\alpha$ are given as \textit{Primary}($\alpha$) = \{$z$ $\vert$ $z$ $\in$ \textit{Secondary}($\alpha$) $\wedge$ $su(\alpha,z)$ $\geq$ \textit{minUtil} $\times $ \textit{TU}\}. The secondary items of $\alpha$ are given as \textit{Secondary}($\alpha$) = \{$z$ $\vert$ $z$ $\in$ $E(\alpha)$ $\wedge$ $lu(\alpha,z)$ $\geq$ \textit{minUtil} $\times$ \textit{TU}\}. Note that \textit{Primary}($\alpha$)	$\subseteq$ \textit{Secondary}($\alpha$).
\end{definition}

Consider the running example and $\alpha$ = \{$A$\}. \textit{Primary($\alpha$)} = $\{B,C\}$, \textit{Secondary($\alpha$)} = $\{B,C,E\}$. In other words, the prefixes $\alpha$ $\cup$ {$B$} and $\alpha$ $\cup$ {$C$} should be explored. Furthermore, only the items \{$B,C,E$\} should be considered as its extensions. To quickly compute the utility of the upper bounds at any time, we use a utility-bin array structure called \textit{UA}. This structure is initialized with 0 before calculation and can be reused, which greatly reduces memory consumption.

\begin{definition}[Utility-bin array \cite{zida2017efim}]
	\rm From database $\mathcal{D}$, there is a set of items $I$. The utility value of the item $i$ is stored in the array \textit{UA}[$i$], and the length of \textit{UA} is $\mid I \mid$.
\end{definition}

\textbf{Using \textit{UA} to calculate \textit{lu}($\alpha$).} First, we initialize the \textit{UA} by filling all elements with 0. Second, the utility-bin array  \textit{UA}[$i$] is updated as \textit{UA}[$i$] = \textit{UA}[$i$] + $u(\alpha,T_c)$ + $re(\alpha,T_c)$, where $i \in T_c \bigcap E(\alpha)$. After database scanning, $\forall i \in E(\alpha)$, \textit{UA}[$i$] = \textit{lu($\alpha$,$i$)}.

\textbf{Using \textit{UA} to calculate \textit{su}($\alpha$).} First, we initialize the \textit{UA} by filling all elements with 0. Second, the utility-bin array \textit{UA}[$i$] is updated as \textit{UA}[$i$] = \textit{UA}[$i$] + $\sum_{I \in T_c \wedge I \in E(\alpha \bigcup i)}u(I,T_c) + u(\alpha,T_c) + u(i,T_c)$, where $i \in T_c \bigcup E(\alpha)$. After database scanning, $\forall i \in E(\alpha)$, \textit{UA}[$i$] = \textit{su($\alpha$,$i$)}.

The above pruning strategies are based on the upper bounds to prune the space. Next, we will introduce the relevant strategies on the correlation.

\begin{Property}
	\label{sec:alg_property4}
	\rm For any itemset $X \in I$, assume $Y$ is an $k$-extension itemset of $X$, where $k \ge 1$. If $X$ is a related pattern, its extension $Y$ is also a related pattern. The $Kulc$ measure is anti-monotonic: \textit{Kulc}($Y$) $\le$ \textit{Kulc}($X$).
\end{Property}

\begin{Strategy}
	\rm Based on Property \ref{sec:alg_property4}, assume that the symbol $\succ$ is sorted in \textit{TWU} ascending order. When utilizing the depth-first search strategy in the search space, if the \textit{Kulc} value of any itemset $X$ is less than the \textit{minCor} value, any of its extension will be regarded as unpromising and irrelevant itemset. They are not CoHUIs which will be ignored. Proof details are provided in the study \cite{gan2018coupm}.    
\end{Strategy}

For example, set \textit{minCor} = 0.7, but the \textit{Kulc} of an itemset $\lbrace A,B \rbrace$ is 0.6. Thus, all its supersets cannot be CoHUIs, and we can remove them from the search space. This anti-monotonic property of \textit{Kulc} is also known as the sorted downward closure property. In other words, if an itemset $X$ is not a strongly related itemset, all its supersets will also not be strongly correlated itemsets.


\subsection{Projection Technology}
\label{sec:alg_projection}
 
The CoIUM algorithm employs a projection technique to reduce the memory overhead. When an itemset $\alpha$ is considered extensible, the algorithm only needs to calculate the utility of $E(\alpha)$. And other items which are not belonged to $E(\alpha)$ will be pruned. The database without these items ($i \notin E(\alpha)$) can be called projected database. Note that all items are sorted based on the \textit{TWU} ascendant order before creating the projected database. The details of the projection mechanism are described below.

\begin{definition}[Projected database \cite{zida2017efim}]
	\rm The $\succ_T$ order is defined as the lexicographical order when reading all transactions from back to frond. Given an itemset $\alpha$ and a database $\mathcal{D}$. The projection of $\alpha$ is denoted as $\alpha$-$\mathcal{D}$, where $\alpha$-$\mathcal{D}$ = \{$\alpha$-$T_c \mid T_c \in \mathcal{D} \wedge \alpha$-$T_c$$\ne$$\emptyset$\}. For any itemset $\alpha$, the projected transaction $T_c$ is denoted as $\alpha$-$T_c$, where $\alpha$-$T_c$ = \{$i \mid i \in T_c \wedge i \in E(\alpha)$\}. The prefix utility of the projected transaction is defined as $pre$($\alpha$-$T_c$) = $u(\alpha,T_c)$.
\end{definition}
 
Let us consider two transactions: $T_a$ = \{$i_1$, $i_2$, $\dots$, $i_m$\} and $T_b$ = \{$j_1$, $j_2$, $\dots$, $j_n$\}, and there are four situations of $\succ_T$. Note that all items are sorted based on \textit{TWU} ascendant order. First, if two transaction have same items, $T_a$ $\succ_T$ $T_b$ when the TID of $T_a$ is greater than that of $T_b$. Second, we compare the last item between $T_a$ and $T_b$. If $T_a$ $\succ_T$ $T_b$, when $m$ $>$ $n$. Third, if the current items are equal after comparing, identify the previous items, when $i_{m-x}$ $\succ$ $j_{n-x}$ (0 $\le$ $x$ $\le$ min($m,n$)), $T_a$ $\succ_T$ $T_b$. Finally, $T_b$ $\succ_T$ $T_a$. 
 
Database projection is a very important technology for reducing database size. As the discovered itemsets become larger, the size of the projected database becomes smaller. The database projection used in CoIUM performs the following. It needs to sort all items based on the \textit{TWU} ascendant order. If it creates the projected transaction of $\alpha$, an offset pointer will be set to $\alpha$ to determine the position of $\alpha$. Then, all transaction containing $\alpha$ will be found, and they are cut into a new projected transaction staring from $\alpha$. Other items which do not belong to the extension of $\alpha$ will not be considered in the new projected transaction.

For example, assume $\alpha$ = $\lbrace E,F \rbrace$. The projected database w.r.t. $\alpha$ is: $\alpha$-$T_2$ = $\lbrace A,B \rbrace$, $\alpha$-$T_7$ = $\lbrace A,D,B,C \rbrace$. The prefix utility of the projected database in corresponding transactions is $pre$(\{$E,F$\}-$T_2$) = $u(\{E,F\},T_2)$ = \$8, $pre$(\{$E,F$\}-$T_7$) = $u(\{E,F\},T_7)$ = \$12. Thus, if the algorithm discovers itemset \{$F,F$\} is a CoHUI, the utility of \{$E,F$\} will be kept in the prefix utility. The prefix utility value of each transaction may be different. When the algorithm expands the itemset \{$E,F$\}, it only needs to select items from the projected database \{$E,F$\}-$\mathcal{D}$ for calculation. Compared to the original database, the size of the projected database is smaller, so it improves the speed of mining CoHUIs. When the algorithm expands the itemset \{$E,F$\}, it only needs to update the value in prefix utility without recalculating the utility of \{$E,F$\}. Thus, using prefix utilities to obtain information can improve the efficiency of algorithm mining CoHUIs.

\subsection{Procedures of CoIUM}
\label{sec:alg_code}

\begin{algorithm}[h]
	\caption{The CoIUM algorithm}
	\label{Algorithm1}
	
		\KwIn{$\mathcal{D}$: transaction database; \textit{minUtil}: a minimum threshold set by user, \textit{minCor}: a positive correlation threshold.}
		\KwOut{a set of CoHUIs.}     
		
	    initialize $\alpha$ = $\emptyset$;
		
		\For{\rm each item $i$ $\in$ \textit{I}}{		
			 scan all transactions, using utility-bin array to compute $lu(\alpha,i)$;
			 calculate the \textit{sup}$(i)$;
		}

		 construct \textit{Secondary($\alpha$)} = \{$i$ $\mid$ $i$ $\in$ \textit{I} $\wedge$ $lu(\alpha,i)$ $\ge$ \textit{minUtil} $\times$ \textit{TU}\};
		
		 update \textit{sup}$(i)$ and \textit{sup}$(i)$ = \{\textit{sup}$(i) \mid i \in$ \textit{Secondary($\alpha$)}\};
		
		\For{\rm each item $i$ $\in$ \textit{I}}{
			 sort \textit{Secondary($\alpha$)} in $\succ$ of \textit{TWU} ascending order;
			
			 eliminate items $i \notin $ \textit{Secondary}$(\alpha) $ from transactions;
			 remove all empty transactions;
			 sort all remaining transactions according to $\succ_T$;
		}

		 construct \textit{Primary($\alpha$)} = \{$i$ $\mid$ $i$ $\in$ \textit{Secondary}$(\alpha)$ $\wedge$ $su(\alpha,i)$ $\ge$ \textit{minUtil} $\times$ \textit{TU}\};

		 call \textbf{Search} ($\alpha$, $\mathcal{D}$, $Primary(\alpha)$, \textit{Secondary($\alpha$)}, \textit{minUtil}, \textit{minCor});
		
		 \textbf{return} CoHUIs
	
\end{algorithm}

\begin{algorithm}[!h]
	\caption{The \textit{Search} procedure}
	\label{sec:Algorithm2}
	\LinesNumbered
	
	   \KwIn{ $\alpha$: the current itemset, $\alpha$-$\mathcal{D}$: projected database w.r.t. $\alpha$, \textit{Primary($\alpha$)}: the primary items of $\alpha$, \textit{Secondary($\alpha$)}: the secondary items of $\alpha$, \textit{minUtil}, \textit{minCor}.} 
		\KwOut{ CoHUIs: a set of CoHUIs.}    
	
	    \For{\rm each \{1\}-length item $i$ $\in$ \textit{Secondary}$(\alpha)$}{
	    \If{u($i$) $\geq$ \textit{minUtil} $\times$ \textit{TU}}{
	    CoHUIs $\leftarrow$ $i$; // the $Kulc$ of $i$ is 1 when $\mid I \mid$ = 1;
    }   
}
	
		\For{\rm each item $i$ $\in$ $Primary(\alpha)$}{
		      $\beta$ = $\alpha$ $\bigcup$ \{$i$\};

			 scan $\alpha$-$\mathcal{D}$ and compute $u(\beta)$;
			\If{\textit{sup($\beta$)} $>$ 0}{
             compute $Kulc(\beta)$;}
			\If {$u(\beta) \geq$ \textit{minUtil} $\times$ \textit{TU} and $Kulc(\beta) \geq$ \textit{minCor}}{ 
			
			 CoHUIs $\leftarrow$ $\beta$;}

				 create $\beta$-$\mathcal{D}$ and scan $\beta$-$\mathcal{D}$ to compute $lu(\beta,z)$ and $su(\beta,z)$, where $\forall z$ $\in$ \textit{Secondary($\alpha$)};
				
				 construct \textit{Primary($\beta$)} = \{$z$ $\in$ \textit{Secondary}($\alpha$) $\vert$ \textit{su}($\beta$,$z$) $\geq$ \textit{minUtil} $\times$ \textit{TU}\};
				
				 construct \textit{Secondary($\beta$)} = \{$z$ $\in$ \textit{Secondary}($\alpha$) $\vert$ \textit{lu}($\beta$,$z$) $\geq$ \textit{minUtil} $\times$ \textit{TU}\};
				
				 call \textbf{Search}\textbf{($\beta$, $\beta$-$\mathcal{D}$,   \textit{Primary($\beta$)},   \textit{Secondary($\beta$)},  \textit{minUtil}, \textit{minCor})};			
	}		
\end{algorithm}

In this subsection, we introduce the details of CoIUM. The main procedure of CoIUM is described in Algorithm \ref{Algorithm1}. Here, three parameters are used as input: two minimum thresholds, \textit{minUtil} and \textit{minCor}, one transaction database $\mathcal{D}$. Firstly, CoIUM defines $\alpha$ as an empty set in line 1. In lines 2-4, it scans the database to compute the support of all items and uses an array-based structure to compute the local utility. Note that if $\alpha$ = $\emptyset$ is taken as the precondition, the value of local utility is equal to the \textit{TWU} value. In line 5, the items whose local utility is greater than or equal to \textit{minUtil} $\times$ \textit{TU} are selected to form the secondary set. In other words, only secondary items will be considered in extensions of $\alpha$. In line 6, the support of items should be updated. In lines 7-10, the algorithm sorts \textit{Secondary($\alpha$)} based on \textit{TWU} ascending order and removes unpromising items that do not belong to \textit{Secondary($\alpha$)}. To reduce the cost of memory, it removes the empty transactions and sorts the remaining transactions based on $\succ_T$. In line 11, CoIUM reuses the array-based structure to compute the subtree utility from the secondary items. If the subtree utility of items is no less than the \textit{minUtil} $\times$ \textit{TU}, they are obtained as primary items. In line 12, the algorithm carries out depth-first exploration from the start of $\alpha$ in a recursive procedure \textit{Search}.

The \textit{Search} procedure (Algorithm \ref{sec:Algorithm2}) takes six parameters as input: the itemset $\alpha$, the projected database w.r.t. $\alpha$, the primary and secondary items w.r.t. $\alpha$, the \textit{minUtil} and \textit{minCor} threshold. In lines 1-5, the algorithm first identifies whether \{l\}-length items are CoHUIs. According to \textit{Kulc}, we can know that the \textit{Kulc} values of \{l\}-length items are always equal to 1. Therefore, the algorithm only needs to compare its own utility value. In lines 6-13, the algorithm starts with item $i$ $\in$ \textit{Primary($\alpha$)}, and these items are considered as extensible items. To obtain the extension of $\alpha$, the algorithm recursively adds items $I$ to form a new itemset $\beta$. Based on the projected database $\alpha$-\textit{D}, the utility of $\beta$ is computed. If the support of itemset $\beta$ is no less than zero, the algorithm should compute its correlation value. Then, if the utility of $\beta$ is greater than \textit{minUtil} $\times$ \textit{TU} and the $Kulc$ value of $\beta$ is greater than the $minCor$, $\beta$ can be output as CoHUIs. In lines 14-16, the algorithm creates a projected database $\beta$-$\mathcal{D}$ and scans this database to compute the local utility and subtree utility w.r.t. $\beta$. The purpose is to find all the extensions of $\beta$ from the secondary items with respect to $\alpha$. Then, it puts the promising items into \textit{Primary}$(\beta)$ and \textit{Secondary}$(\beta)$. In line 17, the \textit{Search} procedure finds the extension of $\beta$ recursively by depth-first method. A complete set of CoHUIs has been output when the CoIUM algorithm terminates.

\subsection{An Example of CoIUM}
\label{sec:alg_example}

In this subsection, we provide a detailed example to discuss how the algorithm works. Assume the database that are displayed in Table \ref{table:database} and Table \ref{table:profit} with \textit{minUtil} = 0.2 and \textit{minCor} = 0.3. In Algorithm \ref{Algorithm1}, it defines $\alpha$ as an empty set (Line 1), then computes the local utility of each item (Line 3). The result of $lu(\alpha,i)$ is that $lu(\alpha,\{A\})$ = \$108, $lu(\alpha,\{B\})$ = \$114, $lu(\alpha,\{C\})$ = \$149, $lu(\alpha,\{D\})$ = \$109, $lu(\alpha,\{E\})$ = \$85, $lu(\alpha,\{F\})$ = \$87. The values of $lu(\alpha,i)$ is equal to \textit{TWU} values when $\alpha$ is empty set. The \textit{minUtil} $\times$ \textit{TU} is (0.2 $\times$ \$167 =) \$33.4, so all the items can be obtained to \textit{Secondary($\alpha$)} and \textit{Secondary($\alpha$)} = \{$A,B,C,D,E,F$\} (Line 5). As shown in Table \ref{table:TWU}, the support of these items are \textit{sup(\{$A$\})} = 5, \textit{sup(\{$B$\})} = 5, \textit{sup(\{$C$\})} = 7, \textit{sup(\{$D$\})} = 5, \textit{sup(\{$E$\})} = 4, and \textit{sup(\{$F$\})} = 4 (Line 3). The support of items can be seen as the number of occurrences of items. Based on \textit{TWU} ascending order $\succ$, \textit{Secondary($\alpha$)} = \{$E,F,A,D,B,C$\} (Line 8). Because no items are removed in \textit{Secondary($\alpha$)}, the database is invariant (Line 9). The transactions are sorted in $\succ_T$ and the results are shown in Table \ref{table:sortTransaction} (Line 9). The result of subtree utility is: $su(\alpha,\{A\})$ = (\$4 + \$4) + (\$4 + \$9 + \$4 + \$2) + (\$16 + \$3) + (\$12 + \$2) + (\$4 + \$9 +\$4 + \$2) = \$79, $su(\alpha,\{B\})$ = \$4 + (\$4 + \$2) + (\$4 + \$3) + (\$4 + \$2) + (\$6 + \$4) = \$33, $su(\alpha,\{C\})$ = \$2 + \$3 + \$2 + \$2 + \$3 + \$2 + \$4 = \$18, $su(\alpha,\{D\})$ = (\$9 + \$4 + \$2) + (\$9 + \$2) + (\$9 + \$4 + \$3) + (\$9 + \$4 + \$2) + \$25 = \$82, $su(\alpha,\{E\})$ = \$18 + \$31 + \$23 + \$13 = \$85, $su(\alpha,\{F\})$ = \$12 + \$21 + \$17 + \$21 = \$71. Thus, the \textit{Primary}($\alpha$) = \{$E,F,A,D$\}, which means that only \{$E,F,A,D$\} will be explored through the depth-first search.

\begin{table}[!htp]
    \normalsize
	\centering
	\caption{Sort transactions and items by $\succ_T$ and $\succ$}
	\setlength{\tabcolsep}{7.4mm}{
	\begin{tabular}{cc} \hline
		\textbf{Tid} &  \textbf{Transaction}    \\ \hline
		$T_2$ & $(E,3)$ $(F,2)$ $(A,1)$ $(B,2) $\\ 
		$T_7$ & $(E,5)$ $(F,2)$ $(A,1)$ $(D,3)$ $(B,2)$ $(C,2) $\\ 	
		$T_3$ & $(E,2)$ $(A,4)$ $(C,3) $\\ 
		$T_8$ & $(E,1)$ $(D,3)$ $(C,2) $\\ 
		$T_4$ & $(F,3)$ $(A,3)$ $(C,2) $\\ 
		$T_6$ & $(F,5)$ $(D,3)$ $(B,2)$ $(C,3) $\\ 
		$T_1$ & $(A,1)$ $(D,3)$ $(B,2)$ $(C,2) $\\ 
		$T_5$ & $(D,5)$ $(B,3)$ $(C,4) $\\ \hline
	\end{tabular}
	}
	\label{table:sortTransaction}
\end{table}

\begin{table}[!htp]
    \normalsize
	\centering
	\caption{The projected database on \{$A$\}}
	\setlength{\tabcolsep}{6.2mm}{
	\begin{tabular}{ccc} \hline
		\textbf{Tid} &  \textbf{Transaction} & \textbf{Prefix utility}  \\ \hline
		$T_2$ & $(B,2) $& \$4\\ 
		$T_7$ & $(D,3)$ $(B,2)$ $(C,2) $ & \$4  \\ 
		$T_3$ & $(C,3) $   &\$16      \\ 
		$T_4$ & $(C,2) $    &\$12      \\ 
		$T_1$ & $(D,3)$ $(B,2)$ $(C,2) $   &\$4         \\ \hline
	\end{tabular}
	}
	\label{table:alg_projected}
\end{table}

In Algorithm \ref{sec:Algorithm2}, it uses the elements of \textit{Primary($\alpha$)} to execute a depth-first search. Because $u(A)$ = \$44, $u(B)$ = \$20, $u(C)$ = \$18, $u(D)$ = \$51, $u(E)$ = \$22, $u(F)$ = \$12, only $A$ and $D$ can be output as CoHUIs. The reason is that $u(\{A\})$ = \$44 ($>$ \textit{minUtil} $\times$ \textit{TU}) and \textit{Kulc(\{$A$\})} = 1 ($>$ 0.3). We assume $\beta$ = \{$A$\} (Line 7), and construct the database projection \{$A$\}-$\mathcal{D}$ in Table \ref{table:alg_projected}. Then, using two utility-bin arrays to compute $lu$ and $su$. The results are: $lu(\beta,\{B\})$ = $tu(T_2)$ + $tu(T_7)$ + $tu(T_1)$ = \$18 + \$31 + \$19 = \$68, $su(\beta,\{B\}) $ = \$8 + \$10 + \$10 = \$28, $lu(\beta,\{C\})$ = $tu(T_7)$ + $tu(T_3)$ + $tu(T_4)$ + $tu(T_1)$ = \$90, $su(\beta,\{C\})$ = \$6 + \$19 + \$14 + \$6 = \$45, $lu(\beta,\{D\})$ = $tu(T_7)$ + $tu(T_1)$ = \$50, and $su(\beta,\{D\})$ = \$38. The primary items and secondary items are updated as: \textit{Primary(\{$\alpha$\})}= \{$D,C$\} and \textit{Secondary(\{$\alpha$\})} = \{$D,B,C$\} (Lines 15--16). This means that only the $\beta$ $\cup$ \{$D,C$\} should be extended, so only items \{$D,B,C$\} can be considered for the depth-first search. Therefore, the algorithm only needs to identify whether itemsets \{$A,D$\}, \{$A,C$\} and their supersets (\{$A,D,B$\}, \{$A,D,C$\}) are CoHUIs.

\section{Experimental Study}
\label{sec:experiments}

In this section, massive experiments were presented to assess the efficiency of CoIUM. Because CoUMP and CoHUI-Miner are the state-of-the-art algorithms for correlated high-utility itemset mining, we selected CoUPM and CoHUI-Miner as the benchmark algorithms. The experiments were performed on a computer with a 3.0 GHz i7-9700 Intel Core Processor with 16 GB of main memory running on Windows 10 (64-bit operating system). All three algorithms involved were implemented by the Java language.

\subsection{Data Description and Experimental Setup}

We evaluated the efficiency of CoIUM algorithm on six spare and dense datasets. All datasets (\textit{BMSPOS}, \textit{Chess}, \textit{Mushroom}, \textit{Retail}, \textit{T10I4D100K} and \textit{T40I10D100K}) are downloaded from the SPMF website$\footnote{SPMF: https://www.philippe-fournier-viger.com/spmf/}$. Table \ref{table:ex_dataset} reveals the detailed features of the experimental datasets. Note that \#Trans represents the quantity of transactions, \#Items represents the summation of different items, \#AvgLen represents the average quantity of items, and \#Type represents two different types: dense and sparse datasets.

\begin{table}[!ht]
	\normalsize
	\centering
	\caption{Experimental dataset characteristics}
	\label{table:ex_dataset}
	\setlength{\tabcolsep}{2.3mm}{
	\begin{tabular}{crrrc} \hline
		\textbf{Dataset}   &   \#\textbf{Trans}    &   \#\textbf{Items} &   \#\textbf{AvgLen} &   \#\textbf{Type}    \\ \hline
		Chess &  3,196   &  75  &  37.0 &  Dense \\ 
		Mushroom &  8,142   &  119  &  23.0 &  Dense \\ 
		T40I10D100K &  100,000   &  942  &  39.6 &  Dense \\ 
		T10I4D100K &  100,000   &  870  &  10.1 &  Sparse \\ 
		Retail &  88,162   &  16,470  &  10.3 &  Sparse \\
		BMSPOS &  515,366   &  1,656  &  6.51 &  Sparse \\ \hline
	\end{tabular}
	}
\end{table}

To assess the efficiency of CoIUM, \textit{minCor} is adjusted twice on each sparse and dense dataset. Consequently, the \textit{minCor} threshold setting is as follows: (1) in \textit{Chess} dataset: 0.2, 0.3; (2) in \textit{Mushroom} dataset: 0.4, 0.5; (3) in \textit{T40I10D100K} dataset: 0.3, 0.4; (4) in \textit{T10I4D100K} dataset: 0.5, 0.6; (5) in \textit{Retail} dataset: 0.6, 0.7; and (6) in \textit{BMSPOS} dataset: 0.1, 0.2.

\begin{figure*}[htbp]
	\centering 
	\includegraphics[trim=50 10 40 10,clip,scale=0.52]{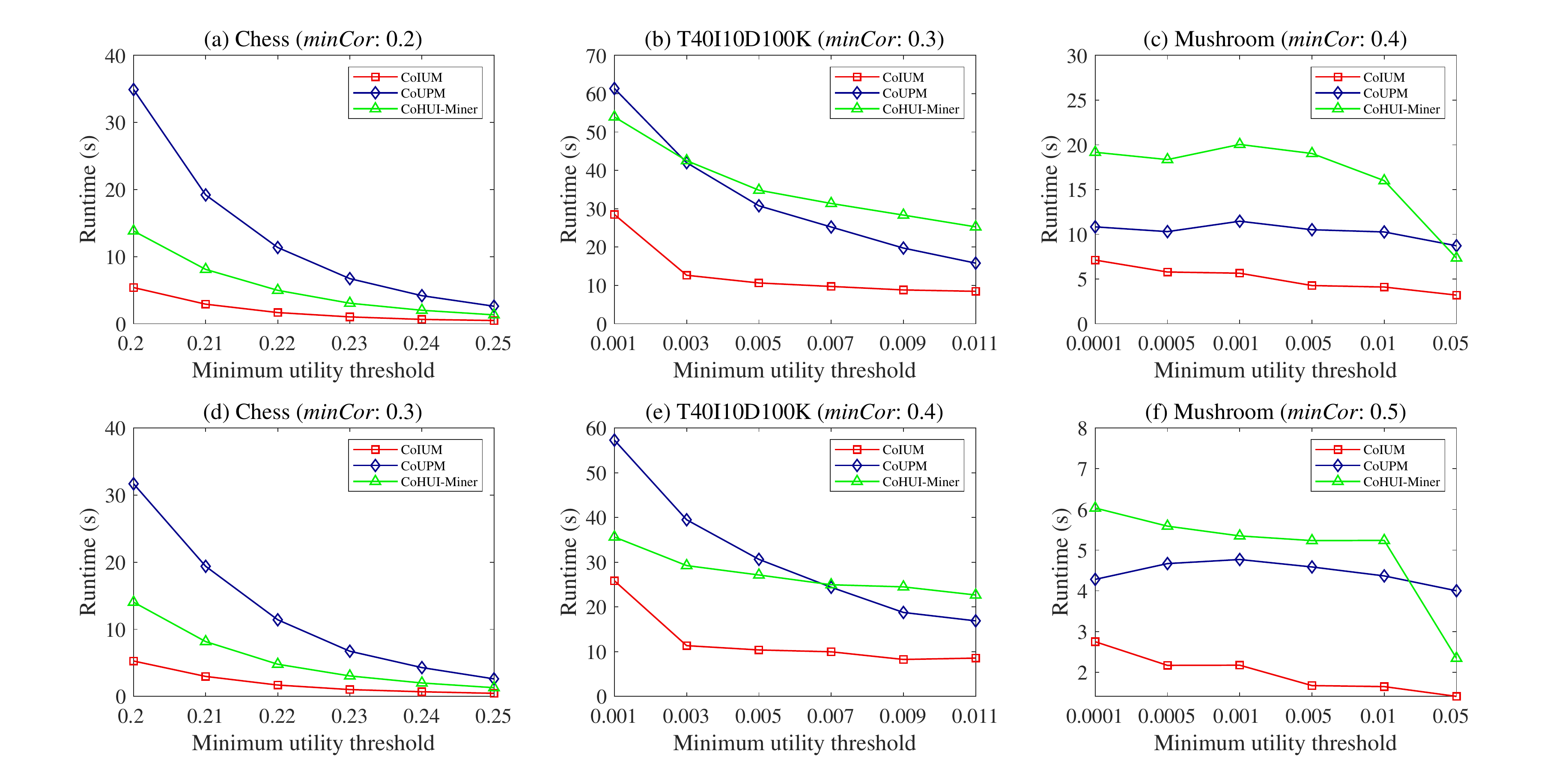}
	\captionsetup{justification=centering}
	\caption{Runtime on dense datasets under changed \textit{minUtil}.}
	\label{fig:Runtime_1}	
\end{figure*}

\begin{figure*}[htbp]
	\centering 
	\includegraphics[trim=40 5 15 20,clip,scale=0.52]{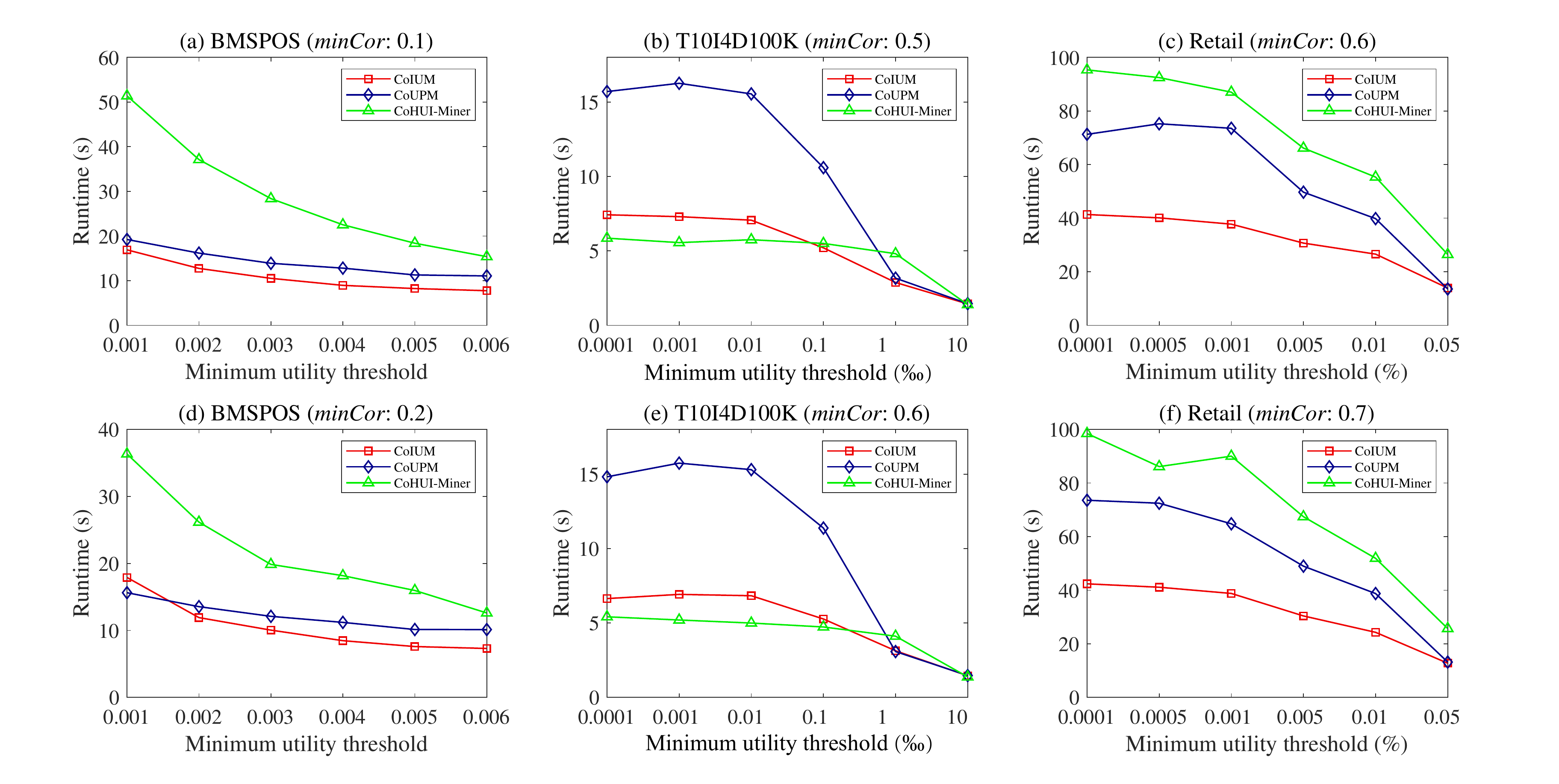}
	\captionsetup{justification=centering}
	\caption{Runtime on sparse datasets under changed \textit{minUtil}.}
	\label{fig:Runtime_2}	
\end{figure*}

\begin{figure*}[h]
	\centering 
	\includegraphics[trim=40 5 15 10,clip,scale=0.52]{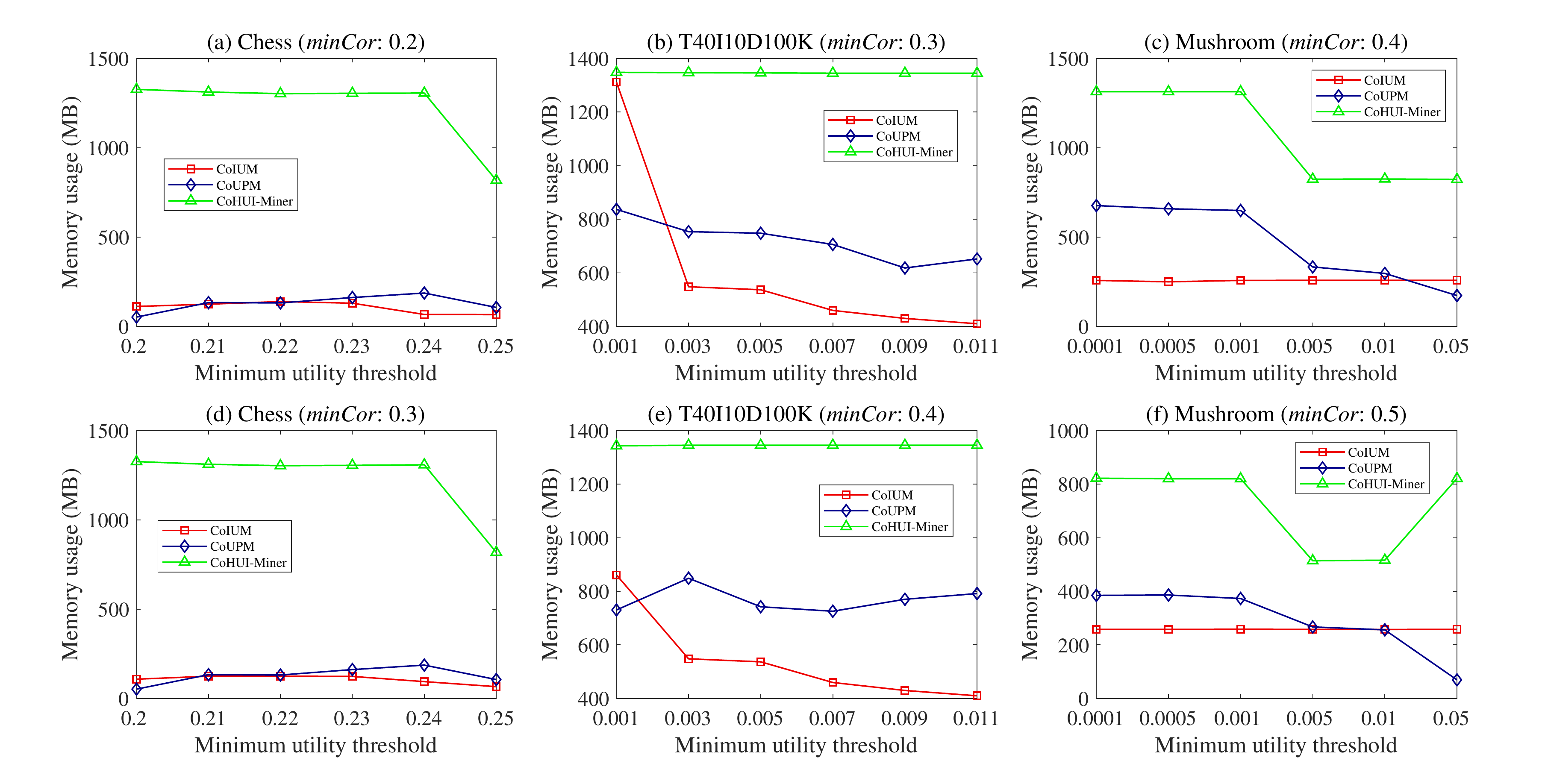}
	\captionsetup{justification=centering}
	\caption{Memory on dense datasets under changed \textit{minUtil}.}
	\label{fig:Memory_1}	
\end{figure*}

\begin{figure*}[h]
	\centering 
	\includegraphics[trim=40 5 15 10,clip,scale=0.52]{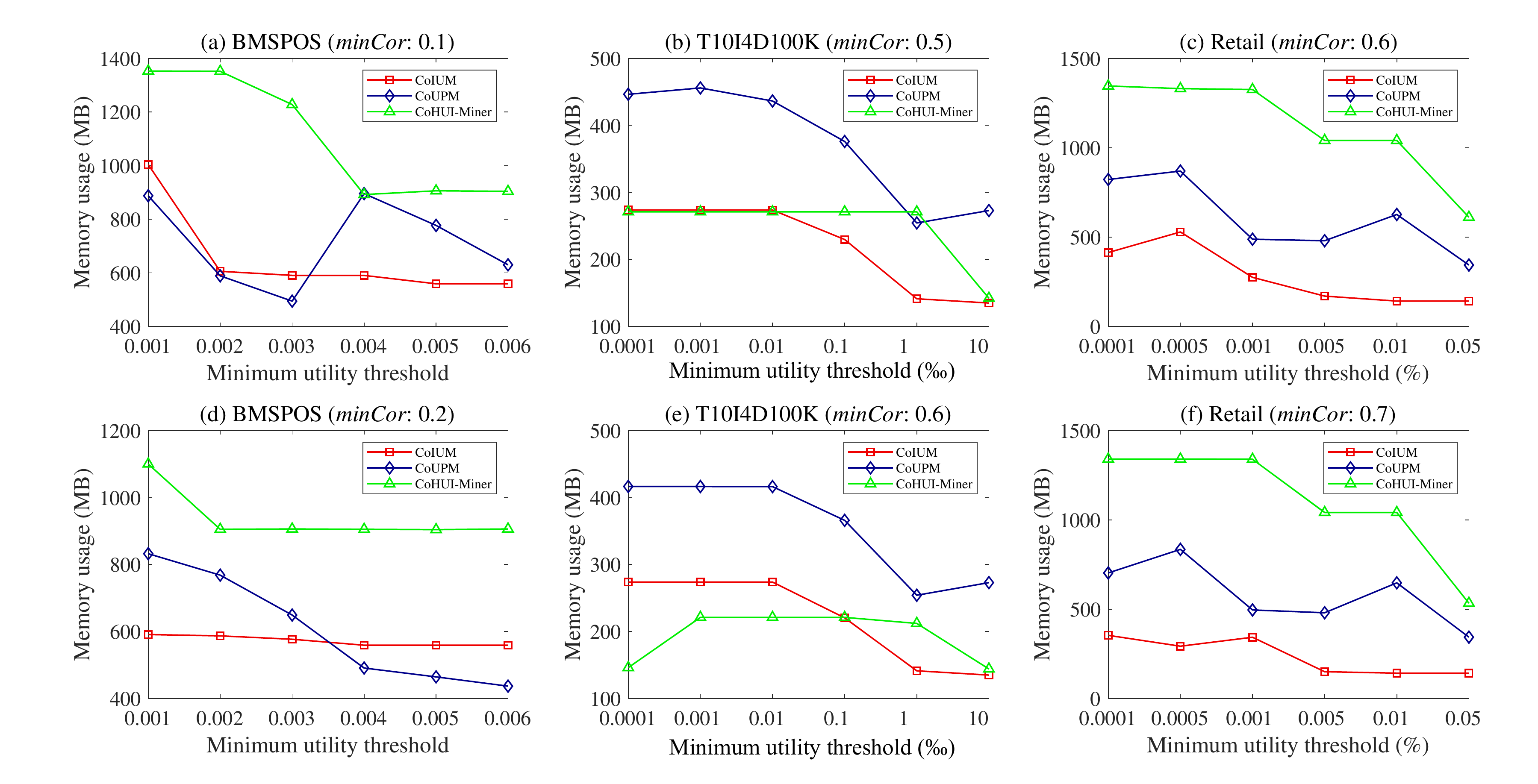}
	\captionsetup{justification=centering}
	\caption{Memory on sparse datasets under changed \textit{minUtil}.}
	\label{fig:Memory_2}	
\end{figure*}

\subsection{Experiments on Runtime}

We compare the runtime of CoIUM with two other algorithms: CoUPM and CoHUI-Miner. The runtime of all three algorithms on dense datasets with various \textit{minCor} values was shown in Fig. \ref{fig:Runtime_1}. The results indicate that CoIUM is faster than CoUPM and CoHUI-Miner on all dense datasets. For example, on the \textit{Chess} dataset with \textit{minCor} = 0.2, the runtime of CoIUM is less than 5 seconds, while that of CoUPM is more than 10 seconds and that of CoHUI-Miner is about 8 seconds. On the \textit{T40I10D100K} dataset with \textit{minCor} = 0.3, the performance of CoIUM is 2X and 3X faster than that of CoUPM and CoHUI-Miner, respectively. On the \textit{Mushroom} dataset with \textit{minCor} = 0.4, the performance of CoIUM is 4X and 2X faster than that of CoHUI-Miner and CoUPM, respectively. We can see that the execution time of CoUPM and CoHUI-Miner increases dramatically as the utility of each dense dataset decreases to a certain point. On the contrary, when the utility of each dense dataset decreases to a certain extent, the execution time of CoIUM remains stable within a certain range.

The runtime of all three algorithms on sparse datasets with different \textit{minUtil} and various \textit{minCor} values was shown in Fig. \ref{fig:Runtime_2}. The results demonstrate that CoIUM is faster than CoUPM and CoHUI-Miner on \textit{BMSPOS} (up to 1.2X and 2.3X, respectively) and \textit{Retail} (up to 1.5X and 2X, respectively). On the \textit{T10I4D100K} dataset, it is interesting that CoIUM has similar performance as CoHUI-Miner and the performance of CoIUM is 3X faster than that of CoUPM. The \textit{minCor} threshold can also affect the efficiency of three algorithms, especially on dense datasets. For example, on the \textit{Mushroom} dataset with \textit{minCor} = 0.4, CoIUM takes about 5 seconds to complete the progress, while CoUPM requires 10 seconds and CoHUI-Miner requires 20 seconds. When \textit{minCor} = 0.5, CoIUM takes about 2 seconds to complete the progress, whereas CoUPM requires 4 seconds and CoHUI-Miner requires 6 seconds. As can be seen, with the increase of \textit{minCor}, the performance of the three algorithms also increases.

The most important reason why CoIUM is often faster than CoUPM and CoHUI-Miner is that CoIUM utilizes two effective upper bounds ($lu$ and $su$). Compared with the other two algorithms, $lu$ and $su$ can prune more space. Therefore, CoIUM only utilizes a few itemsets to discover HUIs. However, CoHUI-Miner has a large size of a projected dataset, which will produce more candidates, especially on dense datasets. CoUPM also takes a lot of time to construct the utility-list structure.

\subsection{Experiments on Memory Usage}

Memory consumption is an important indicator for evaluating the merits of data mining algorithms, so in this subsection, we will compare the memory usage between CoIUM, CoUPM, and CoHUI-Miner. Based on the same variable setting as shown in Fig. \ref{fig:Runtime_1}, the detailed results on the dense datasets are shown in Fig. \ref{fig:Memory_1}. It can be clearly obtained that the memory consumption of CoIUM is much less than that of CoHUI-Miner on all dense datasets. For example, on the \textit{T40I10D100K} dataset with \textit{minCor} = 0.3, the memory loss of CoHUI-Miner is always maintained at about 1400 MB, while the memory usage of CoIUM is less than 1000 MB and remains below 500 MB. This result proves that the performance of CoIUM is much better than that of CoHUI-Miner, because CoHUI-Miner produces too many candidates and requires a lot of memory to store information. Then, on the \textit{Chess} dataset, the memory usage of CoIUM is nearly the same but lower than CoUPM. However, the memory usage of CoIUM is often lower than that of CoUPM on \textit{T40I10D100K} and \textit{Mushroom} datasets. For instance, on the \textit{Mushroom} dataset with \textit{minCor} = 0.4, the memory usage of CoIUM tends to remain at 220 MB, while CoUPM remains between 200 MB and 600 MB. In most cases, the memory consumption of CoUPM exceeds 220 MB.

Based on the same variable setting as displayed in Fig. \ref{fig:Runtime_2}, the detailed results on the sparse datasets are shown in Fig. \ref{fig:Memory_2}. As we can observe from the \textit{Retail} dataset, the memory usage of CoIUM is much lower than that of both CoUPM and CoHUI-Miner (less than 2X and 3X, respectively). On the \textit{BMSPOS} and \textit{T10I4D100K} datasets, the memory usage of CoIUM is lower than that of CoHUI-Miner, and close to that of CoUPM when \textit{minUtil} increases. Because of sparse datasets, the size of the transaction dataset is small, so both CoUPM and CoHUI-Miner require only a small amount of memory to conserve candidates.

The main reason why the memory consumption of CoIUM is lower than that of CoUPM in all datasets is that the projected technique can reduce dataset size more effectively than the utility-list used in the CoUPM algorithm. Although both CoIUM and CoHUI-Miner use the same projected technique, the consumption of CoIUM is lower than CoHUI-Miner on all datasets. The reason is that both of them use different strategies to explore CoHUIs. CoIUM uses two upper bounds ($lu$ and $su$) to reduce the number of candidates generated. These two upper bounds are more compact than \textit{TWU} utilized in CoHUI-Miner, so they can remove more unpromising candidates than \textit{TWU}. The number of candidates produced by CoIUM is lower than that of CoUPM and CoHUI-Miner. As a result, the memory consumption of CoIUM is lower than that of CoUPM and CoHUI-Miner.

\begin{table*}[!h]
	\fontsize{8pt}{10pt}\selectfont
	\centering
	\caption{\# Candidate generation under varying \textit{minUtil} with fixed \textit{minCor}}
	\label{table:patterns1}
	\setlength{\tabcolsep}{3mm}{
		\begin{tabular}{ccrrrrrrr}
			\hline\hline
			\multirow{2}*{\textbf{Dataset}}&	\multirow{2}*{\textbf{\textit{minCor}}}&
			\multirow{2}*{\textbf{\# Algorithm}}
			&\multicolumn{6}{c}{\textbf{\# candidate generation under different thresholds}}\\
			\cline{4-9}
			&&&$ \delta_1 $ & $ \delta_2 $ & $ \delta_3 $ & $ \delta_4 $ &  $ \delta_5 $  & $ \delta_6 $ \\ \hline
			
			&  &  \textbf{\#CoIUM}  &	59,039  &	28,384  &	13,808  &	6,614  &	3,187  &	1,501	 \\
			
			&0.2 &  \textbf{\#CoUPM}  &	 200,606  &	111,453  &	63,757  &	37,323  &	22,593  &	13,794 	 \\
			& &\textbf{\#CoHUI-Miner}  &	 2620,117  &	1240,899  &	595,812  &	288,709  &	140,696  &	67,589	 \\
			
			\cline{2-9}
			\textbf{Chess}
			&  &  \textbf{\#CoIUM} &	59,039  &	28,384  &	13,808  &	6,614  &	3,187  &	1,501	 \\
			
			&0.3 &  \textbf{\#CoUPM}   &	 200,606  &	111,453  &	63,757  &	37,323  &	22,593  &	13,794 	 \\
			& &\textbf{\#CoHUI-Miner}  &	 2,620,117  &	1,240,899  &	595,812  &	288,709  &	140,696  &	67,589	 \\
			
			\hline
			
			&  &  \textbf{\#CoIUM} &	152,930  &	15,935  &	3,738  &	1,475  &	957  &	796	 \\
			
			&0.3 &  \textbf{\#CoUPM}   &	 447,807  &	140,222  &	69,670  &	35,506  &	21,179  &	13,068 	 \\
			& &\textbf{\#CoHUI-Miner}  &	 17435,132  &	7761,519 &	3077,564  &	1362,147  &	863,037  &	626,828	 \\
			
			\cline{2-9}
			\textbf{T40I10D100K}
			&  &  \textbf{\#CoIUM} &	112,925  &	14,147  &	3,549  &	1,468  &	957  &	796	 \\
			
			&0.4 &  \textbf{\#CoUPM}   &	 245,896  &	95,410  &52,987  &	31,469  &	19,752  &	12,724 	 \\
			& &\textbf{\#CoHUI-Miner}  &	 2449,760  &	1563,543 &	1107,732  &	823,317  &	673,769  &	585,371	 \\
			
			\hline
			&  &  \textbf{\#CoIUM} &	276,518  &	269,858  &	253,590  &	97,812  &	88,277  &	45,928	 \\
			
			&0.4 &  \textbf{\#CoUPM}   &	 281,139  &	277,600  &267,469  &	105,995  &	96,669  &	57,134 	 \\
			& &\textbf{\#CoHUI-Miner}  &	 13,045,179  &	12,697,922 &	12,367,803  &	4,615,310  &	4,083,339  &	2,015,229	 \\
			
			\cline{2-9}
			\textbf{Mushroom}
			&  &  \textbf{\#CoIUM}
			&	98,887  &	96,816  &	88,517  
			&	33,401  &	30,094  &	17,187	 \\
			&0.5 &  \textbf{\#CoUPM}   
			&	 102,875  &	101,615  &97,757  
			&	38,210  &	34,701  &	22,059 	 \\
			& &\textbf{\#CoHUI-Miner}  
			&	 3,170,082  &	3,093,407 &	3,004,022  
			&	1,073,346  &	952,846  &	525,577	 \\		
			\hline
			
			&  &  \textbf{\#CoIUM}
			&	3,747,571  &	3,420,817  &	2,659,589  
			&	233,859  &	48,056  &	6,779	 \\
			&0.6 &  \textbf{\#CoUPM}   
			&	 134,722,599  &	129,039,485  &120,942,326  
			&	73,168,998  &	49,245,865  &	10,939,731	 \\
			& &\textbf{\#CoHUI-Miner}  
			&	 533,955,947  &	485,002,701 &	389,635,998  
			&	157,096,263  &	107,204,879  &	30,288,360	 \\		
			\cline{2-9}
			\textbf{Retail}	
			
			&  &  \textbf{\#CoIUM}
			&	3,615,594  &	3,297,971  &	2,570,798  
			&	231,801  &	47,899  &	6,773	 \\
			&0.7 &  \textbf{\#CoUPM}   
			&	 134,588,149  &	128,906,771  &120,821,509  
			&	73,158,370  &	49,245,259  &	10,939,665	 \\
			& &\textbf{\#CoHUI-Miner}  
			&	 311,399,670  &	292,452,873 &	264,220,651  
			&	154,067,657  &	106,915,555  &	30,200,504	 \\		
			\hline		
			
			\hline
			\hline
		\end{tabular}
	}
\end{table*}

\subsection{Experiments on Candidates Generation}

Table \ref{table:patterns1} shows the candidate generation results of CoIUM, CoUPM, and CoHUI-Miner, under different \textit{minUtil} and \textit{minCor}. Note that $ \delta $ represents a different \textit{minUtil} threshold, and its specific scope is from Fig. \ref{fig:Runtime_1}(a) to (c) and Fig. \ref{fig:Runtime_2}(a) to (c). The results indicate that the candidate generation by CoIUM is lower than CoUPM and CoHUI-Miner on dense datasets. For example, on \textit{T40I10D100K} dataset with \textit{minCor} = 0.3, the quantity of candidates increased by CoUPM is more than 2X higher than that of CoIUM, and the quantity of candidates generated by CoHUI-Miner is up to 10 orders of magnitude higher than that of CoIUM. With the increase of \textit{minUtil}, the decrease of CoIUM is significantly greater than that of CoUPM and CoHUI-Miner. On the \textit{Mushroom} dataset with \textit{minCor} = 0.4, the quantity of candidates increased by CoIUM is always close to but lower than that of CoUPM, and the quantity of candidates produced by CoHUI-Miner is up to four orders of magnitude higher than that of CoIUM. These results prove that both CoUPM and CoHUI-Miner discover numerous meaningless Itemsets, which is why the runtime of CoIUM is better than CoUPM and CoHUI-Miner on all dense datasets. The reason why CoIUM generated fewer candidates than CoUPM and CoHUI-Miner on dense datasets is that the CoIUM algorithm adopts two tighter upper bounds (\textit{lu} and \textit{su}). These upper bounds can be used to remove more unpromising itemsets that do not satisfy the properties of CoHUIs.

On a spare dataset, it is found that the sum of candidates produced by CoIUM is also lower than that of CoUPM and CoHUI-Miner, especially on \textit{Retail} dataset. For example, on \textit{Retail} dataset with \textit{minCor} = 0.6, the sum of candidates generated by CoUPM and CoHUI-Miner is up to 4 and 10 orders of magnitude higher than CoIUM respectively. And with the increase of \textit{minUtil}, the gap between the number of candidates they produce becomes larger. With the increase of \textit{minCor} on different datasets except \textit{Chess}, the number of candidates generated by the three algorithms decreases correspondingly. It can be seen that \textit{minCor} also has an impact on the discovery of HUIs. Because the data of \textit{Chess} dataset is too dense, the small promotion of \textit{minCor} cannot effectively filter out unpromising candidates.

\subsection{Experiments on Scalability Test}

\begin{figure}[h]
	\centering 
	\includegraphics[trim=25 10 15 10,clip,scale=0.5]{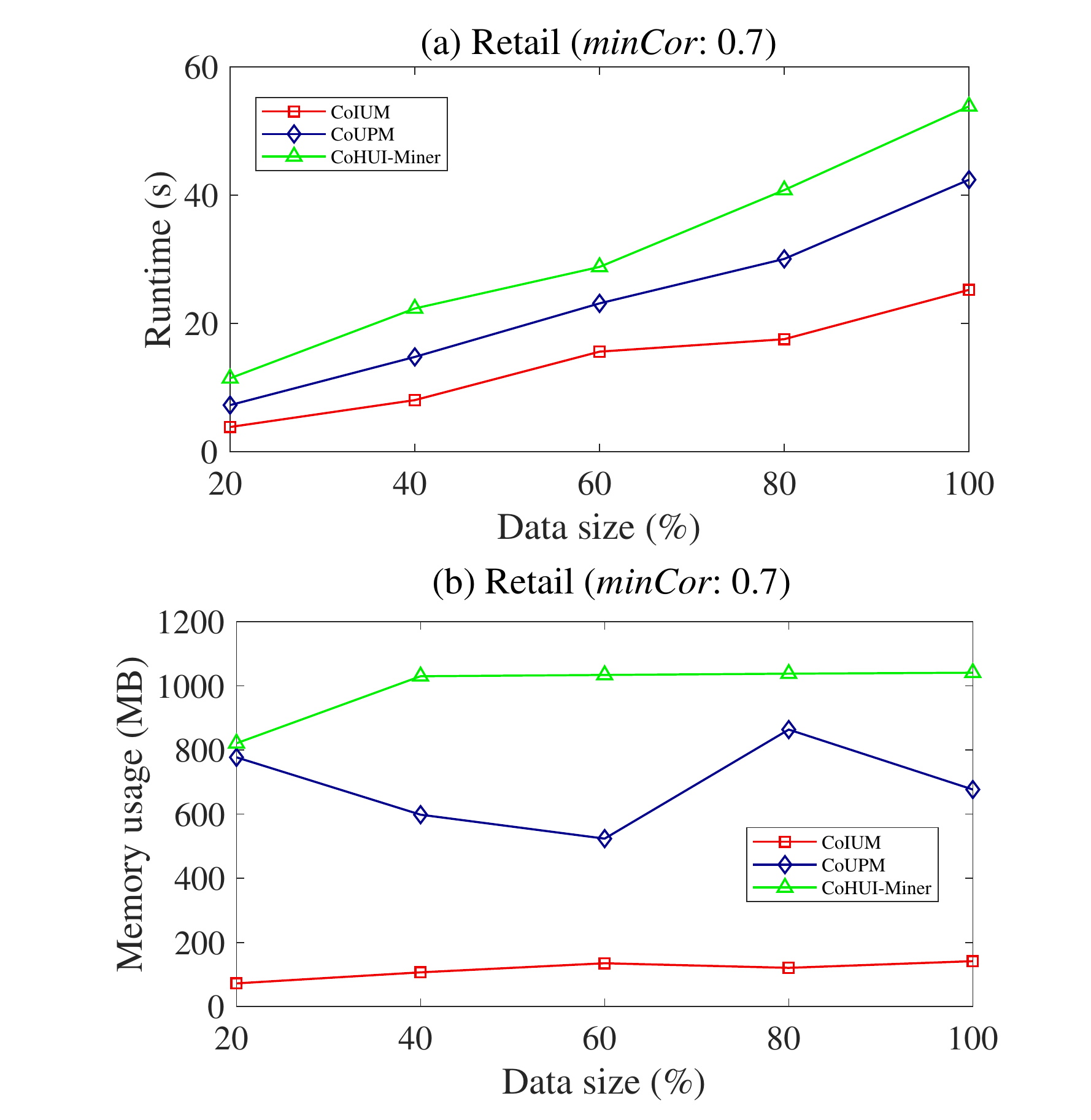}
	\captionsetup{justification=centering}
	\caption{Runtime and memory scalability on \textit{Retail}.}
	\label{fig:scability}	
\end{figure}

Moreover, we test the scalability of CoIUM, CoUPM, and CoHUI-Miner. We adopt the size of the \textit{Retail} dataset from 20\% (= 17,632 transactions) to 100\% (= 88,162 transactions) with \textit{minCor} = 0.7. Fig. \ref{fig:scability} shows the details of execution time and memory usage. From Fig. \ref{fig:scability}(a), We can know that the runtime of CoIUM, CoUPM, and CoHUI-Miner respectively increases linearly with the increasing dataset size. Compared with the other two algorithms, CoIUM has a smaller runtime increment because \textit{lu} and \textit{su} can reduce more unpromising candidates than that of \textit{TWU}. At the same time, the gap in runtime between CoIUM and the other two algorithms is becoming larger. This conclusion proves that the pruning strategies of CoIUM are obviously better than those of the other two algorithms. In addition, we can see from Fig. \ref{fig:scability}(b) that with the growth of dataset size, the memory usage of CoIUM increases slowly and remains at about 100 MB, and the memory use of CoHUI-Miner also increases slowly and remains at about 1000 MB after a significant increase, while the memory use of CoUPM remains a decrease-increase-decrease phenomenon. We can see that with the increase in dataset size, the memory trend of CoIUM is more gentle than that of the other two algorithms. The results demonstrate that CoIUM has preferable scalability in runtime and memory consumption compared with CoUPM and CoHUI-Miner. In addition, the execution time and memory usage of CoIUM increase slowly when the data size increases. These results indicate that CoIUM is more suitable for large-scale datasets.

\section{Conclusions} \label{sec:conclusion}

In this study, we present an efficient algorithm named CoIUM to extract the set of CoHUIs. The proposed CoIUM algorithm adopts two new upper bounds called local utility and subtree utility, which can prune more unpromising candidates effectively. Then, a novel array-based structure known as \textit{UA} is applied to store useful information when pruning the search space. A novel technique called database projection is designed to reduce the cost of datasets in memory. In addition, we utilize \textit{TWU} ascending order to sort all items. Extensive experiments on different datasets show that the performance of the CoIUM algorithm is obviously better than the CoUPM and CoHUI-Miner algorithms.

In the future, we would like to explore other novel data structures to reduce memory consumption effectively. We will extend the designed algorithm to different fields such as on-shelf utility mining, HUIM without setting \textit{minUtil} and fuzzy utility mining. We also plan to apply the proposed algorithm to resolve other types of complex data (e.g., episodes, streaming data), not just static data.




\bibliographystyle{IEEEtran}
\bibliography{coium}


\end{document}